\RequirePackage{booktabs}
\documentclass[runningheads]{llncs}
\usepackage[T1]{fontenc}
\usepackage{amsmath}
\usepackage{multirow}
\usepackage{subcaption}
\usepackage[table,xcdraw,dvipsnames]{xcolor}
\definecolor{MyGray}{HTML}{F5F5F5}
\definecolor{MyOrange}{HTML}{FFCE93}
\definecolor{MyYellow}{HTML}{FCFCCB}
\definecolor{MyRed}{HTML}{FCE5CD}
\usepackage{graphicx}%

\usepackage[normalem]{ulem}
\useunder{\uline}{\ul}{}

\usepackage{amsmath,amssymb,amsfonts}%

\usepackage{mathrsfs}%
\usepackage[title]{appendix}%
\usepackage{textcomp}%

\usepackage{booktabs}%
\usepackage{manyfoot}%

\usepackage{algorithm}%
\usepackage{algorithmicx}%
\usepackage{algpseudocode}%
\usepackage{listings}
\usepackage{lipsum}  
\usepackage{lscape}

\usepackage{orcidlink} 
\usepackage{hyperref}

\makeatletter
\newcommand\notsotiny{\@setfontsize\notsotiny{6.31415}{9}}
\makeatother

\makeatletter
\renewcommand*{\@fnsymbol}[1]{\ensuremath{\ifcase#1\or *\or \dagger\or \ddagger\or
   \mathsection\or \mathparagraph\or \|\or **\or \dagger\dagger
   \or \ddagger\ddagger \else\@ctrerr\fi}}
\makeatother

\begin{document}
\title{Comics Datasets Framework: Mix of Comics datasets for detection benchmarking}
\titlerunning{Comics Datasets Framework}
%
\author{Emanuele Vivoli\thanks{Corresponding author: \href{mailto:evivoli@cvc.uab.cat}{evivoli@cvc.uab.cat}}\thanks{Annotation team}\inst{1,2}\orcidlink{0000-0002-9971-8738} \and
Irene Campaioli$^{\dag}$\inst{2}\orcidlink{0009-0002-4231-1686} \and
Mariateresa Nardoni$^{\dag}$\inst{2}\orcidlink{0009-0002-8528-7890} \and \\
Niccolò Biondi\inst{2}\orcidlink{0000-0003-1153-1651} \and
Marco Bertini\inst{2}\orcidlink{0000-0002-1364-218X} \and
Dimosthenis Karatzas\inst{1}\orcidlink{0000-0001-8762-4454}}
\authorrunning{E. Vivoli et al.}
%
\institute{Computer Vision Center, UAB, Spain \and
MICC, University of Florence, Italy}
\maketitle              
\begin{abstract}
Comics, as a medium, uniquely combine text and images in styles often distinct from real-world visuals. For the past three decades, computational research on comics has evolved from basic object detection to more sophisticated tasks. However, the field faces persistent challenges such as small datasets, inconsistent annotations, inaccessible model weights, and results that cannot be directly compared due to varying train/test splits and metrics. To address these issues, we aim to standardize annotations across datasets, introduce a variety of comic styles into the datasets, and establish benchmark results with clear, replicable settings. Our proposed Comics Datasets Framework standardizes dataset annotations into a common format and addresses the overrepresentation of manga by introducing Comics100, a curated collection of 100 books from the Digital Comics Museum, annotated for detection in our uniform format. We have benchmarked a variety of detection architectures using the Comics Datasets Framework. All related code, model weights, and detailed evaluation processes are available at \href{https://github.com/emanuelevivoli/cdf}{https://github.com/emanuelevivoli/cdf}, ensuring transparency and facilitating replication. This initiative is a significant advancement towards improving object detection in comics, laying the groundwork for more complex computational tasks dependent on precise object recognition.

\end{abstract}
\section{Introduction}
Comics represent a unique form of media that integrates text with graphical elements. This medium has gained widespread popularity as a form of cultural expression globally, from ``American Comics'' in the US, to ``Bandes Dessinées'' in France and Belgium and ``Manga'' in Japan. Despite their apparent simplicity and accessibility, even for children, the intricate layouts of comic pages present substantial computational challenges. The classic components such as panels, balloons, characters, text, and onomatopoeia are highly influenced by the author's creativity and artistic style, making the task of comic image analysis a non-trivial endeavor.

In recent works \cite{li_manga109dialog_2023,sachdeva_manga_2024}, authors approached Comics research to more complex tasks starting from \textit{detection} to \textit{comics dialog generation}. However, many of the more difficult tasks are based on the simplest one, i.e. generating the dialog for a comics \cite{li_manga109dialog_2023} requires the model to recognize the elements, link the text to the characters, identify multiple instances of the same character and sort textboxes within panels and the panels themself. One mistake in this process will propagate to the next stages of the pipeline making an erroneous final dialog generation. 

Despite these ``classical'' computer vision tasks being well-explored in other domains, they are far from being solved in comics. This challenge can be attributed primarily to two factors: (i) datasets' annotations and size, and (ii) model availability. If we look at comics field, the available datasets are relatively small in size and available annotations do not always agree. In fact, the biggest annotated dataset is Manga109 \cite{fujimoto_manga109_2016} which provide panels, characters, text, face, onomatopoeias and dialog for the 10.6k images available. Another very famous dataset is DCM \cite{nguyen_digital_2018}, which instead has 772 pages, fully annotated and with comics style. Lastly, eBDtheque \cite{guerin_ebdtheque_2013} provide similar annotations for just 100 pages of mixed styles (Japanese, American and French comics). The second issue, regarding model availability, mainly impact the reproducibility of experiments. Due to this second factor, little progress is possible and new works cannot compare with previous state-of-the-art models due to non-standard train/test splits, and the unavailability of models.

In this work we aim to solve these two issues: we collected four of the main available datasets, and provided unified annotations of common objects, fixing the missing ones. As the majority of them are Manga-styled images, we annotated 100 American comic books resulting in more than 5k images. Finally, we benchmark available models for detection, and fine-tuned three CNNs model classes (namely Faster R-CNN, SSD and YOLO). We provide the research community with the code to manage the dataset's images \footnote{for copyright issues we cannot share the datasets' images}, the corrected annotations and to evaluate against CoMix, the collection of more than 29k comic book pages of multiple styles and languages.

\section{Related works}

\paragraph{\textbf{Comics Datasets.}}
The field of comics analysis includes various datasets aimed at processing and studying comic media. However, many of these datasets are either no longer available or require specific permissions from research groups to use. This kind of restriction is common with copyrighted materials, which are generally available only for research purposes. If redistribution is prohibited, it is usual to offer annotations along with a script that facilitates the downloading and organization of the dataset, as exemplified in \cite{sachdeva_manga_2024}. 
The datasets vary greatly in quality and size. For instance, the dataset \textit{Manga109} \cite{fujimoto_manga109_2016} is a well-annotated and modest-size manga dataset (109 books, 10.6k images of double-sided pages), with recent works expanding it to include annotations for onomatopoeias \cite{baek_coo_2022} and dialogue \cite{li_manga109dialog_2023}. In contrast, the \textit{COMICS} dataset by Nguyen et al. \cite{nguyen_comic_2017} is larger (3.9k books, 198k pages) but has limited, automatically generated labels and has seen only minimal updates in recent years \cite{agrawal_multimodal_2023,vivoli_multimodal_2024}. Other highly-curated but very small datasets are \textit{eBDtheque} \cite{guerin_ebdtheque_2013}, a collection of 100 pages over 20 books, designed for detection and segmentation, and \textit{DCM} \cite{nguyen_digital_2018}, collection of 772 pages from Digital Comic Museum\footnote{\href{https://digitalcomicmuseum.com}{www.digitalcomicmuseum.com}} containing panels, characters and faces bounding boxes. Lastly, a recent dataset of English Manga-style comics has been proposed in \cite{sachdeva_manga_2024} named PopManga, containing almost 2k test split with detection annotations and text-character and character-character links for dialog transcription.
The datasets discussed primarily consist of comics in English, French, and Japanese, and notably lack recent publications (post-2010)\footnote{Only PopManga contains a few chapters of 2020 volumes.} due to copyright restrictions. The creation of private, small-scale datasets and the challenges in replicating studies are significant issues in the comic domain. In our work, we tackle these issues by focusing on enhancing the accessibility of datasets and facilitating the sharing of models.

\paragraph{\textbf{Detection Models in Comics.}}
Object detection is a prevalent task in comic analysis. It was first introduced in \cite{guerin_ebdtheque_2013} (2013), but recently appears to be one of the main task in \cite{dutta_bcbid_2022} (2022) and \cite{sachdeva_manga_2024} (2024). In Table \ref{tab:previous-benchmarks}, we present the highest performance metrics by task and dataset, citing the model type and the original works.
However, comparisons of state-of-the-art (SOTA) performances are often not straightforward: access to code is frequently unavailable, and the only models for which weights are accessible are the 2022 DASS \cite{topal_dassdetector_2023} and 2024 Magi \cite{sachdeva_manga_2024}, leaving many model architecture un-replicable.
Moreover, other missing information such as training/testing splits and metric calculation methods, makes any comparison unfair. As highlighted in \cite{padilla_comparative_2021}, significant differences exist among standard detection algorithms and their metric calculations. Keeping this in mind, we have selected the model families for our detection tasks (Faster R-CNN, SSD and YOLO) and benchmarked them uniformly. The evaluation script is part of our contribution to the \textit{CDF} repository.

\begin{table}[]
\centering
\tiny
\sffamily
\begin{tabular}{lcccccc}
\rowcolor[HTML]{FFFFFF} 
\multicolumn{2}{c}{\cellcolor[HTML]{FFFFFF}\textbf{DATASETS INFO}} & \multicolumn{5}{c}{\cellcolor[HTML]{FFFFFF}\textbf{BENCHMARKS}} \\
\rowcolor[HTML]{FFFFFF} 
\multicolumn{1}{c}{} &  &  &  & \multicolumn{1}{c}{} &  &  \\ 
\rowcolor[HTML]{FFFFFF} 
\multicolumn{1}{c}{\cellcolor[HTML]{FFFFFF}\textbf{name}} & \textbf{format} & \textbf{task} & \textbf{model type} & \multicolumn{1}{c}{\cellcolor[HTML]{FFFFFF}\textbf{work}} & \textbf{metric} & \textbf{perf (\%)} \\ 
\rowcolor[HTML]{FFFFFF} 
\multicolumn{1}{c}{} &  &  &  & \multicolumn{1}{c}{} &  &  \\ 
\rowcolor[HTML]{FFFFFF} 
\multicolumn{1}{l}{\cellcolor[HTML]{FFFFFF}\textbf{Fahad18 \cite{khan_color_2012}}} & - & det [C] & \textbf{-} & \multicolumn{1}{c}{\cellcolor[HTML]{FFFFFF}\cite{khan_color_2012}} & mAP & 41,7 \\ 
\rowcolor[HTML]{EFEFEF} 
\multicolumn{1}{l}{\cellcolor[HTML]{EFEFEF}} & \cellcolor[HTML]{EFEFEF} & \cellcolor[HTML]{EFEFEF} & \cellcolor[HTML]{EFEFEF} & \multicolumn{1}{c}{\cellcolor[HTML]{EFEFEF}} & P & 91,5 \\ \cline{6-7} 
\rowcolor[HTML]{EFEFEF} 
\multicolumn{1}{l}{\cellcolor[HTML]{EFEFEF}} & \cellcolor[HTML]{EFEFEF} & \multirow{-2}{*}{\cellcolor[HTML]{EFEFEF}gm} & \multirow{-2}{*}{\cellcolor[HTML]{EFEFEF}\textbf{graph}} & \multicolumn{1}{c}{\multirow{-2}{*}{\cellcolor[HTML]{EFEFEF}\cite{ho_redundant_2013}}} & R & 71,5 \\ \cline{3-7} 
\rowcolor[HTML]{EFEFEF} 
\multicolumn{1}{l}{\multirow{-3}{*}{\cellcolor[HTML]{EFEFEF}\textbf{Ho42 \cite{ho_redundant_2013}}}} & \multirow{-3}{*}{\cellcolor[HTML]{EFEFEF}-} & det [C] & \textbf{graph} & \multicolumn{1}{c}{\cellcolor[HTML]{EFEFEF}\cite{ho_redundant_2013}} & Acc & 71,4 \\ 
\rowcolor[HTML]{FFFFFF} 
\multicolumn{1}{l}{\cellcolor[HTML]{FFFFFF}} & \cellcolor[HTML]{FFFFFF} & \cellcolor[HTML]{FFFFFF} & \cellcolor[HTML]{FFFFFF} & \multicolumn{1}{c}{\cellcolor[HTML]{FFFFFF}} & P & 93,5 \\ \cline{6-7} 
\rowcolor[HTML]{FFFFFF} 
\multicolumn{1}{l}{\cellcolor[HTML]{FFFFFF}} & \cellcolor[HTML]{FFFFFF} & \cellcolor[HTML]{FFFFFF} & \cellcolor[HTML]{FFFFFF} & \multicolumn{1}{c}{\cellcolor[HTML]{FFFFFF}} & R & 96,2 \\ \cline{6-7} 
\rowcolor[HTML]{FFFFFF} 
\multicolumn{1}{l}{\cellcolor[HTML]{FFFFFF}} & \cellcolor[HTML]{FFFFFF} & \multirow{-3}{*}{\cellcolor[HTML]{FFFFFF}seg [B]} & \multirow{-3}{*}{\cellcolor[HTML]{FFFFFF}\textbf{custom- CNN}} & \multicolumn{1}{c}{\multirow{-3}{*}{\cellcolor[HTML]{FFFFFF}\cite{dutta_cnnbased_2020}}} & F1 & 94,8 \\ \cline{3-7} 
\rowcolor[HTML]{FFFFFF} 
\multicolumn{1}{l}{\cellcolor[HTML]{FFFFFF}} & \cellcolor[HTML]{FFFFFF} & \cellcolor[HTML]{FFFFFF} & \cellcolor[HTML]{FFFFFF} & \multicolumn{1}{c}{\cellcolor[HTML]{FFFFFF}} & P & 75,2 \\ \cline{6-7} 
\rowcolor[HTML]{FFFFFF} 
\multicolumn{1}{l}{\cellcolor[HTML]{FFFFFF}} & \cellcolor[HTML]{FFFFFF} & \cellcolor[HTML]{FFFFFF} & \cellcolor[HTML]{FFFFFF} & \multicolumn{1}{c}{\cellcolor[HTML]{FFFFFF}} & R & 49,8 \\ \cline{6-7} 
\rowcolor[HTML]{FFFFFF} 
\multicolumn{1}{l}{\multirow{-6}{*}{\cellcolor[HTML]{FFFFFF}\textbf{eBDtheque \cite{guerin_ebdtheque_2013}}}} & \multirow{-6}{*}{\cellcolor[HTML]{FFFFFF}SVG/XML} & \multirow{-3}{*}{\cellcolor[HTML]{FFFFFF}det [F]} & \multirow{-3}{*}{\cellcolor[HTML]{FFFFFF}\textbf{Faster R-CNN}} & \multicolumn{1}{c}{\multirow{-3}{*}{\cellcolor[HTML]{FFFFFF}\cite{qin_faster_2017}}} & F1 & 60 \\ 
\rowcolor[HTML]{EFEFEF} 
\multicolumn{1}{l}{\cellcolor[HTML]{EFEFEF}} & \cellcolor[HTML]{EFEFEF} & \cellcolor[HTML]{EFEFEF} & \cellcolor[HTML]{EFEFEF} & \multicolumn{1}{c}{\cellcolor[HTML]{EFEFEF}} & P & 97,8 \\
\rowcolor[HTML]{EFEFEF} 
\multicolumn{1}{l}{\multirow{-2}{*}{\cellcolor[HTML]{EFEFEF}\textbf{sun70 \cite{sun_specific_2013}}}} & \multirow{-2}{*}{\cellcolor[HTML]{EFEFEF}-} & \multirow{-2}{*}{\cellcolor[HTML]{EFEFEF}det [C]} & \multirow{-2}{*}{\cellcolor[HTML]{EFEFEF}\textbf{SIFT}} & \multicolumn{1}{c}{\multirow{-2}{*}{\cellcolor[HTML]{EFEFEF}\cite{sun_specific_2013}}} & R & 47 \\ 
\rowcolor[HTML]{FFFFFF} 
\multicolumn{1}{l}{\cellcolor[HTML]{FFFFFF}} & \cellcolor[HTML]{FFFFFF} & \cellcolor[HTML]{FFFFFF} & \cellcolor[HTML]{FFFFFF} & \multicolumn{1}{c}{\cellcolor[HTML]{FFFFFF}} & P & 75,4 \\ \cline{6-7} 
\rowcolor[HTML]{FFFFFF} 
\multicolumn{1}{l}{\cellcolor[HTML]{FFFFFF}} & \cellcolor[HTML]{FFFFFF} & \cellcolor[HTML]{FFFFFF} & \cellcolor[HTML]{FFFFFF} & \multicolumn{1}{c}{\cellcolor[HTML]{FFFFFF}} & R & 9,8 \\ \cline{6-7} 
\rowcolor[HTML]{FFFFFF} 
\multicolumn{1}{l}{\cellcolor[HTML]{FFFFFF}} & \cellcolor[HTML]{FFFFFF} & \cellcolor[HTML]{FFFFFF} & \cellcolor[HTML]{FFFFFF} & \multicolumn{1}{c}{\cellcolor[HTML]{FFFFFF}} & ScoreP & 82,18 \\ \cline{6-7} 
\rowcolor[HTML]{FFFFFF} 
\multicolumn{1}{l}{\multirow{-4}{*}{\cellcolor[HTML]{FFFFFF}\textbf{SSGCI \cite{le_subgraph_2018}}}} & \multirow{-4}{*}{\cellcolor[HTML]{FFFFFF}XML} & \multirow{-4}{*}{\cellcolor[HTML]{FFFFFF}subg-s} & \multirow{-4}{*}{\cellcolor[HTML]{FFFFFF}\textbf{graph}} & \multicolumn{1}{c}{\multirow{-4}{*}{\cellcolor[HTML]{FFFFFF}\cite{le_subgraph_2018}}} & ScoreR & 80,71 \\ 
\rowcolor[HTML]{EFEFEF} 
\multicolumn{1}{l}{\cellcolor[HTML]{EFEFEF}} & \cellcolor[HTML]{EFEFEF} & T-c [easy] & \textbf{ComicVT5} & \multicolumn{1}{c}{\cellcolor[HTML]{EFEFEF}\cite{vivoli_multimodal_2024}} & Acc & 79,1 \\ \cline{3-7} 
\rowcolor[HTML]{EFEFEF} 
\multicolumn{1}{l}{\cellcolor[HTML]{EFEFEF}} & \cellcolor[HTML]{EFEFEF} & T-c [hard] & \textbf{ComicVT5} & \multicolumn{1}{c}{\cellcolor[HTML]{EFEFEF}\cite{vivoli_multimodal_2024}} & Acc & 71,3 \\ \cline{3-7} 
\rowcolor[HTML]{EFEFEF} 
\multicolumn{1}{l}{\cellcolor[HTML]{EFEFEF}} & \cellcolor[HTML]{EFEFEF} & V-c [easy] & \textbf{CNN + LSTM} & \multicolumn{1}{c}{\cellcolor[HTML]{EFEFEF}\cite{iyyer_amazing_2017}} & Acc & 85,7 \\ \cline{3-7} 
\rowcolor[HTML]{EFEFEF} 
\multicolumn{1}{l}{\cellcolor[HTML]{EFEFEF}} & \cellcolor[HTML]{EFEFEF} & V-c [hard] & \textbf{CNN + LSTM} & \multicolumn{1}{c}{\cellcolor[HTML]{EFEFEF}\cite{iyyer_amazing_2017}} & Acc & 63,2 \\ \cline{3-7} 
\rowcolor[HTML]{EFEFEF} 
\multicolumn{1}{l}{\multirow{-5}{*}{\cellcolor[HTML]{EFEFEF}\textbf{COMICS \cite{iyyer_amazing_2017}}}} & \multirow{-5}{*}{\cellcolor[HTML]{EFEFEF}TXT} & C-c & \textbf{CNN + LSTM} & \multicolumn{1}{c}{\cellcolor[HTML]{EFEFEF}\cite{iyyer_amazing_2017}} & Acc & 70,9 \\ 
\rowcolor[HTML]{FFFFFF} 
\multicolumn{1}{l}{\cellcolor[HTML]{FFFFFF}} & \cellcolor[HTML]{FFFFFF} & \cellcolor[HTML]{FFFFFF} & \cellcolor[HTML]{FFFFFF} & \multicolumn{1}{c}{\cellcolor[HTML]{FFFFFF}} & P & 99,24 \\ \cline{6-7} 
\rowcolor[HTML]{FFFFFF} 
\multicolumn{1}{l}{\cellcolor[HTML]{FFFFFF}} & \cellcolor[HTML]{FFFFFF} & \cellcolor[HTML]{FFFFFF} & \cellcolor[HTML]{FFFFFF} & \multicolumn{1}{c}{\cellcolor[HTML]{FFFFFF}} & R & 99,16 \\ \cline{6-7} 
\rowcolor[HTML]{FFFFFF} 
\multicolumn{1}{l}{\multirow{-3}{*}{\cellcolor[HTML]{FFFFFF}\textbf{Comics3w \cite{he_sren_2017}}}} & \multirow{-3}{*}{\cellcolor[HTML]{FFFFFF}-} & \multirow{-3}{*}{\cellcolor[HTML]{FFFFFF}det [P]} & \multirow{-3}{*}{\cellcolor[HTML]{FFFFFF}custom- Faster R-CNN} & \multicolumn{1}{c}{\multirow{-3}{*}{\cellcolor[HTML]{FFFFFF}\cite{he_endtoend_2018}}} & F1 & 99,2 \\ 
\rowcolor[HTML]{EFEFEF} 
\multicolumn{1}{l}{\cellcolor[HTML]{EFEFEF}} & \cellcolor[HTML]{EFEFEF} & \cellcolor[HTML]{EFEFEF} & \cellcolor[HTML]{EFEFEF} & \multicolumn{1}{c}{\cellcolor[HTML]{EFEFEF}} & P & 95 \\ \cline{6-7} 
\rowcolor[HTML]{EFEFEF} 
\multicolumn{1}{l}{\cellcolor[HTML]{EFEFEF}} & \cellcolor[HTML]{EFEFEF} & \cellcolor[HTML]{EFEFEF} & \cellcolor[HTML]{EFEFEF} & \multicolumn{1}{c}{\cellcolor[HTML]{EFEFEF}} & R & 93,2 \\ \cline{6-7} 
\rowcolor[HTML]{EFEFEF} 
\multicolumn{1}{l}{\multirow{-3}{*}{\cellcolor[HTML]{EFEFEF}\textbf{JC2463 \cite{qin_faster_2017}}}} & \multirow{-3}{*}{\cellcolor[HTML]{EFEFEF}-} & \multirow{-3}{*}{\cellcolor[HTML]{EFEFEF}det [F]} & \multirow{-3}{*}{\cellcolor[HTML]{EFEFEF}\textbf{Faster R-CNN}} & \multicolumn{1}{c}{\multirow{-3}{*}{\cellcolor[HTML]{EFEFEF}\cite{qin_faster_2017}}} & F1 & 94,1 \\ 
\rowcolor[HTML]{FFFFFF} 
\multicolumn{1}{l}{\cellcolor[HTML]{FFFFFF}} & \cellcolor[HTML]{FFFFFF} & \cellcolor[HTML]{FFFFFF} & \cellcolor[HTML]{FFFFFF} & \multicolumn{1}{c}{\cellcolor[HTML]{FFFFFF}} & P & 82,4 \\ \cline{6-7} 
\rowcolor[HTML]{FFFFFF} 
\multicolumn{1}{l}{\cellcolor[HTML]{FFFFFF}} & \cellcolor[HTML]{FFFFFF} & \cellcolor[HTML]{FFFFFF} & \cellcolor[HTML]{FFFFFF} & \multicolumn{1}{c}{\cellcolor[HTML]{FFFFFF}} & R & 73,1 \\ \cline{6-7} 
\rowcolor[HTML]{FFFFFF} 
\multicolumn{1}{l}{\multirow{-3}{*}{\cellcolor[HTML]{FFFFFF}\textbf{AEC912 \cite{qin_faster_2017}}}} & \multirow{-3}{*}{\cellcolor[HTML]{FFFFFF}-} & \multirow{-3}{*}{\cellcolor[HTML]{FFFFFF}det [F]} & \multirow{-3}{*}{\cellcolor[HTML]{FFFFFF}\textbf{Faster R-CNN}} & \multicolumn{1}{c}{\multirow{-3}{*}{\cellcolor[HTML]{FFFFFF}\cite{qin_faster_2017}}} & F1 & 77,5 \\ 
\rowcolor[HTML]{EFEFEF} 
\multicolumn{1}{l}{\cellcolor[HTML]{EFEFEF}} & \cellcolor[HTML]{EFEFEF} & \cellcolor[HTML]{EFEFEF} & \cellcolor[HTML]{EFEFEF} & \multicolumn{1}{c}{\cellcolor[HTML]{EFEFEF}} & P & 95,58 \\ \cline{6-7} 
\rowcolor[HTML]{EFEFEF} 
\multicolumn{1}{l}{\cellcolor[HTML]{EFEFEF}} & \cellcolor[HTML]{EFEFEF} & \cellcolor[HTML]{EFEFEF} & \cellcolor[HTML]{EFEFEF} & \multicolumn{1}{c}{\cellcolor[HTML]{EFEFEF}} & R & 94,04 \\ \cline{6-7} 
\rowcolor[HTML]{EFEFEF} 
\multicolumn{1}{l}{\multirow{-3}{*}{\cellcolor[HTML]{EFEFEF}\textbf{GNC \cite{dunst_graphic_2017}}}} & \multirow{-3}{*}{\cellcolor[HTML]{EFEFEF}CSV} & \multirow{-3}{*}{\cellcolor[HTML]{EFEFEF}det [B]} & \multirow{-3}{*}{\cellcolor[HTML]{EFEFEF}\textbf{U-Net (VGG-16)}} & \multicolumn{1}{c}{\multirow{-3}{*}{\cellcolor[HTML]{EFEFEF}\cite{dubray_deep_2019}}} & F1 & 94,48 \\ 
\rowcolor[HTML]{FFFFFF} 
\multicolumn{1}{l}{\cellcolor[HTML]{FFFFFF}} & \cellcolor[HTML]{FFFFFF} & \cellcolor[HTML]{FFFFFF} & \cellcolor[HTML]{FFFFFF} & \multicolumn{1}{c}{\cellcolor[HTML]{FFFFFF}} & P & 92,7 \\ \cline{6-7} 
\rowcolor[HTML]{FFFFFF} 
\multicolumn{1}{l}{\cellcolor[HTML]{FFFFFF}} & \cellcolor[HTML]{FFFFFF} & \cellcolor[HTML]{FFFFFF} & \cellcolor[HTML]{FFFFFF} & \multicolumn{1}{c}{\cellcolor[HTML]{FFFFFF}} & R & 96,9 \\ \cline{6-7} 
\rowcolor[HTML]{FFFFFF} 
\multicolumn{1}{l}{\multirow{-3}{*}{\cellcolor[HTML]{FFFFFF}\textbf{DCM772 \cite{nguyen_digital_2018}}}} & \multirow{-3}{*}{\cellcolor[HTML]{FFFFFF}TXT} & \multirow{-3}{*}{\cellcolor[HTML]{FFFFFF}det [T]} & \multirow{-3}{*}{\cellcolor[HTML]{FFFFFF}custom- CNN} & \multicolumn{1}{c}{\multirow{-3}{*}{\cellcolor[HTML]{FFFFFF}\cite{dutta_cnnbased_2020}}} & F1 & 94,7 \\ 
\rowcolor[HTML]{EFEFEF} 
\multicolumn{1}{l}{\cellcolor[HTML]{EFEFEF}} & \cellcolor[HTML]{EFEFEF} & \cellcolor[HTML]{EFEFEF} & \cellcolor[HTML]{EFEFEF} & \multicolumn{1}{c}{\cellcolor[HTML]{EFEFEF}} & P & 93,56 \\ \cline{6-7} 
\rowcolor[HTML]{EFEFEF} 
\multicolumn{1}{l}{\cellcolor[HTML]{EFEFEF}} & \cellcolor[HTML]{EFEFEF} & \cellcolor[HTML]{EFEFEF} & \cellcolor[HTML]{EFEFEF} & \multicolumn{1}{c}{\cellcolor[HTML]{EFEFEF}} & R & 95,49 \\ \cline{6-7} 
\rowcolor[HTML]{EFEFEF} 
\multicolumn{1}{l}{\cellcolor[HTML]{EFEFEF}} & \cellcolor[HTML]{EFEFEF} & \multirow{-3}{*}{\cellcolor[HTML]{EFEFEF}seg [B]} & \multirow{-3}{*}{\cellcolor[HTML]{EFEFEF}custom- CNN} & \multicolumn{1}{c}{\multirow{-3}{*}{\cellcolor[HTML]{EFEFEF}\cite{dutta_cnnbased_2020}}} & F1 & 94,51 \\ \cline{3-7} 
\rowcolor[HTML]{EFEFEF} 
\multicolumn{1}{l}{\cellcolor[HTML]{EFEFEF}} & \cellcolor[HTML]{EFEFEF} & det [P] & \textbf{SSD300} & \multicolumn{1}{c}{\cellcolor[HTML]{EFEFEF}\cite{ogawa_object_2018}} & Acc & 97,1 \\ \cline{3-7} 
\rowcolor[HTML]{EFEFEF} 
\multicolumn{1}{l}{\cellcolor[HTML]{EFEFEF}} & \cellcolor[HTML]{EFEFEF} & det [T] & custom- SSD300 & \multicolumn{1}{c}{\cellcolor[HTML]{EFEFEF}\cite{ogawa_object_2018}} & Acc & 84.1 \\ \cline{3-7} 
\rowcolor[HTML]{EFEFEF} 
\multicolumn{1}{l}{\cellcolor[HTML]{EFEFEF}} & \cellcolor[HTML]{EFEFEF} & det [F] & custom- SSD300 & \multicolumn{1}{c}{\cellcolor[HTML]{EFEFEF}\cite{ogawa_object_2018}} & Acc & 76.2 \\ \cline{3-7} 
\rowcolor[HTML]{EFEFEF} 
\multicolumn{1}{l}{\multirow{-7}{*}{\cellcolor[HTML]{EFEFEF}\textbf{Manga109-anns \cite{fujimoto_manga109_2016,ogawa_object_2018}}}} & \multirow{-7}{*}{\cellcolor[HTML]{EFEFEF}XML} & det [C] & custom- SSD300 & \multicolumn{1}{c}{\cellcolor[HTML]{EFEFEF}\cite{ogawa_object_2018}} & Acc & 79.6 \\ 
\rowcolor[HTML]{FFFFFF} 
\multicolumn{1}{l}{\cellcolor[HTML]{FFFFFF}} & \cellcolor[HTML]{FFFFFF} & \cellcolor[HTML]{FFFFFF} & \cellcolor[HTML]{FFFFFF} & \multicolumn{1}{c}{\cellcolor[HTML]{FFFFFF}} & P & 91,01 \\ \cline{6-7} 
\rowcolor[HTML]{FFFFFF} 
\multicolumn{1}{l}{\cellcolor[HTML]{FFFFFF}} & \cellcolor[HTML]{FFFFFF} & \cellcolor[HTML]{FFFFFF} & \cellcolor[HTML]{FFFFFF} & \multicolumn{1}{c}{\cellcolor[HTML]{FFFFFF}} & R & 91,23 \\ \cline{6-7} 
\rowcolor[HTML]{FFFFFF} 
\multicolumn{1}{l}{\multirow{-3}{*}{\cellcolor[HTML]{FFFFFF}\textbf{Sequencity4k \cite{nguyen_learning_2020}}}} & \multirow{-3}{*}{\cellcolor[HTML]{FFFFFF}-} & \multirow{-3}{*}{\cellcolor[HTML]{FFFFFF}seg [B]} & \multirow{-3}{*}{\cellcolor[HTML]{FFFFFF}\textbf{U-Net (VGG-16)}} & \multicolumn{1}{c}{\multirow{-3}{*}{\cellcolor[HTML]{FFFFFF}\cite{nguyen_learning_2020}}} & F1 & 91.12 (+- 5.44) \\ 
\rowcolor[HTML]{EFEFEF} 
\multicolumn{1}{l}{\cellcolor[HTML]{EFEFEF}\textbf{EmoRecCom \cite{nguyen_icdar_2021}}} & CSV & cls [E] & \textbf{CNN + BERT} & \multicolumn{1}{c}{\cellcolor[HTML]{EFEFEF}\cite{nguyen_icdar_2021}} & AUC & 68,49 \\ 
\rowcolor[HTML]{FFFFFF} 
\multicolumn{1}{l}{\cellcolor[HTML]{FFFFFF}} & \cellcolor[HTML]{FFFFFF} & \cellcolor[HTML]{FFFFFF} & \cellcolor[HTML]{FFFFFF} & \multicolumn{1}{c}{\cellcolor[HTML]{FFFFFF}} & P & 97,05 \\ \cline{6-7} 
\rowcolor[HTML]{FFFFFF} 
\multicolumn{1}{l}{\cellcolor[HTML]{FFFFFF}} & \cellcolor[HTML]{FFFFFF} & \cellcolor[HTML]{FFFFFF} & \cellcolor[HTML]{FFFFFF} & \multicolumn{1}{c}{\cellcolor[HTML]{FFFFFF}} & R & 98,81 \\ \cline{6-7} 
\rowcolor[HTML]{FFFFFF} 
\multicolumn{1}{l}{\cellcolor[HTML]{FFFFFF}} & \cellcolor[HTML]{FFFFFF} & \multirow{-3}{*}{\cellcolor[HTML]{FFFFFF}seg [B]} & \multirow{-3}{*}{\cellcolor[HTML]{FFFFFF}custom- CNN} & \multicolumn{1}{c}{\multirow{-3}{*}{\cellcolor[HTML]{FFFFFF}\cite{dutta_cnnbased_2020}}} & F1 & 97,92 \\ \cline{3-7} 
\rowcolor[HTML]{FFFFFF} 
\multicolumn{1}{l}{\cellcolor[HTML]{FFFFFF}} & \cellcolor[HTML]{FFFFFF} & \cellcolor[HTML]{FFFFFF} & \cellcolor[HTML]{FFFFFF} & \multicolumn{1}{c}{\cellcolor[HTML]{FFFFFF}} & P & 95,63 \\ \cline{6-7} 
\rowcolor[HTML]{FFFFFF} 
\multicolumn{1}{l}{\cellcolor[HTML]{FFFFFF}} & \cellcolor[HTML]{FFFFFF} & \cellcolor[HTML]{FFFFFF} & \cellcolor[HTML]{FFFFFF} & \multicolumn{1}{c}{\cellcolor[HTML]{FFFFFF}} & R & 98,52 \\ \cline{6-7} 
\rowcolor[HTML]{FFFFFF} 
\multicolumn{1}{l}{\multirow{-6}{*}{\cellcolor[HTML]{FFFFFF}\textbf{BCBId (Bangla) \cite{dutta_bcbid_2022}}}} & \multirow{-6}{*}{\cellcolor[HTML]{FFFFFF}TXT/XML} & \multirow{-3}{*}{\cellcolor[HTML]{FFFFFF}det [T]} & \multirow{-3}{*}{\cellcolor[HTML]{FFFFFF}custom- CNN} & \multicolumn{1}{c}{\multirow{-3}{*}{\cellcolor[HTML]{FFFFFF}\cite{dutta_cnnbased_2020}}} & F1 & 97,05 \\ 
\rowcolor[HTML]{EFEFEF} 
\multicolumn{1}{l}{\cellcolor[HTML]{EFEFEF}} & \cellcolor[HTML]{EFEFEF} & \cellcolor[HTML]{EFEFEF} & \cellcolor[HTML]{EFEFEF} & \multicolumn{1}{c}{\cellcolor[HTML]{EFEFEF}} & P & 69,8 \\ \cline{6-7} 
\rowcolor[HTML]{EFEFEF} 
\multicolumn{1}{l}{\cellcolor[HTML]{EFEFEF}} & \cellcolor[HTML]{EFEFEF} & \cellcolor[HTML]{EFEFEF} & \cellcolor[HTML]{EFEFEF} & \multicolumn{1}{c}{\cellcolor[HTML]{EFEFEF}} & R & 65,9 \\ \cline{6-7} 
\rowcolor[HTML]{EFEFEF} 
\multicolumn{1}{l}{\cellcolor[HTML]{EFEFEF}} & \cellcolor[HTML]{EFEFEF} & \multirow{-3}{*}{\cellcolor[HTML]{EFEFEF}det [O]} & \multirow{-3}{*}{\cellcolor[HTML]{EFEFEF}\textbf{MTSv3}} & \multicolumn{1}{c}{\multirow{-3}{*}{\cellcolor[HTML]{EFEFEF}\cite{baek_coo_2022}}} & H & 67,8 \\ \cline{3-7} 
\rowcolor[HTML]{EFEFEF} 
\multicolumn{1}{l}{\cellcolor[HTML]{EFEFEF}} & \cellcolor[HTML]{EFEFEF} & rec [T] & \textbf{CNN + BiLSTM} & \multicolumn{1}{c}{\cellcolor[HTML]{EFEFEF}\cite{baek_coo_2022}} & Acc & 81 \\ \cline{3-7} 
\rowcolor[HTML]{EFEFEF} 
\multicolumn{1}{l}{\cellcolor[HTML]{EFEFEF}} & \cellcolor[HTML]{EFEFEF} & \cellcolor[HTML]{EFEFEF} & \cellcolor[HTML]{EFEFEF} & \multicolumn{1}{c}{\cellcolor[HTML]{EFEFEF}} & P & 77,2 \\ \cline{6-7} 
\rowcolor[HTML]{EFEFEF} 
\multicolumn{1}{l}{\cellcolor[HTML]{EFEFEF}} & \cellcolor[HTML]{EFEFEF} & \cellcolor[HTML]{EFEFEF} & \cellcolor[HTML]{EFEFEF} & \multicolumn{1}{c}{\cellcolor[HTML]{EFEFEF}} & R & 68,7 \\ \cline{6-7} 
\rowcolor[HTML]{EFEFEF} 
\multicolumn{1}{l}{\multirow{-7}{*}{\cellcolor[HTML]{EFEFEF}\textbf{COO \cite{baek_coo_2022}}}} & \multirow{-7}{*}{\cellcolor[HTML]{EFEFEF}XML} & \multirow{-3}{*}{\cellcolor[HTML]{EFEFEF}l-p [O]} & \multirow{-3}{*}{\cellcolor[HTML]{EFEFEF}\textbf{custom- M4C}} & \multicolumn{1}{c}{\multirow{-3}{*}{\cellcolor[HTML]{EFEFEF}\cite{baek_coo_2022}}} & H & 72,7 \\ 
\end{tabular}
\vspace{5mm}
\caption{Overview of existing Benchmarks. \textit{Only the highest values are reported.} Abbreviations represent [tasks]: Detection (det), Segmentation (seg), graph matching (gm), subgraph spotting (subg-s), Text-cloze (T-c), Visual-cloze (T-c), Character-coherence (C-c), classification (cls), recognition (rec), link-prediction (l-p). These tasks can be applied to different objects: Character [C], balloon [B], face [F], panel [P], text [T], emotion [E], and onomatopoeia [O]. Metrics are: mean Average Precision (mAP), Precision (P), Recall (R), and Accuracy (Acc).}
\label{tab:previous-benchmarks}
\end{table}

\section{Comics Datasets Framework}
In this section, we describe the structure of the \textit{CDF} repository (Sec. \ref{meth:structure}), the unified annotations format (Sec. \ref{meth:uca-format}), the supported datasets (Sec. \ref{meth:supp-datasets}), and the evaluation procedure (Sec. \ref{meth:evaluation}).

\subsection{Structure}
\label{meth:structure}
\begin{figure}
    \centering
    \includegraphics[width=\textwidth]{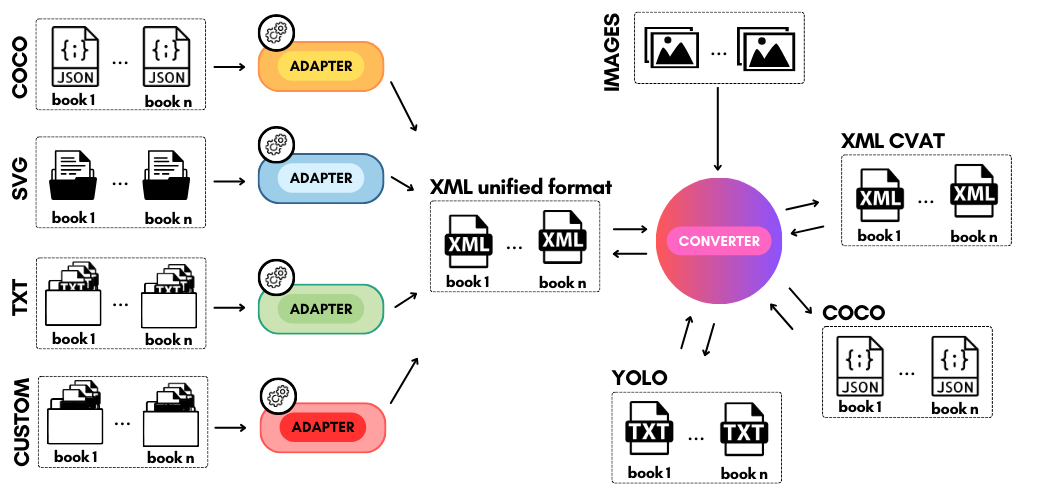}
    \caption{Unification pipeline schema: given a dataset in an origin format, through a specialized adapter we obtain the XML unified format. This can be converted to CVAT, COCO, or any format required.}
    \label{fig:structure}
\end{figure}

The two crucial stages of our pipeline include (i) adapting datasets from any format to our standardized structure, and (ii) transforming this unified structure into widely recognized formats such as CVAT, COCO, and others. The ability to support multiple conversions is essential for specific applications: the CVAT standard is utilized by the corresponding annotation tool, which is instrumental for handling and updating existing annotations; the COCO standard is widely adopted for computer vision annotations and predictions, extensively used for evaluation with cocotools\footnote{\href{https://cocodataset.org/\#detection-eval}{https://cocodataset.org/detection-eval}}. As shown in Figure \ref{fig:structure}, our annotation conversion pipeline necessitates a distinct adapter for each dataset structure received, and a converter for each target format supported.

\subsection{UCA format}
\label{meth:uca-format}
In an effort to standardize the analysis of comic book datasets, we introduce the \textit{Unified Comics Annotation} (UCA) format. UCA is an XML-based format, inspired by Manga109 \cite{fujimoto_manga109_2016}, that systematically categorizes various elements commonly found in comics. The primary goal of this format is to facilitate detailed and structured annotations that support a broad spectrum of comic analysis tasks. 
The core structure of UCA is hierarchical, beginning with the root element \textit{<book>}, which encompasses all subsequent data. This encapsulates metadata such as the book title, and sub-elements that detail \textit{<characters>}, \textit{<stories>}, and \textit{<pages>}.
Each page within a book is represented by a \textit{<page>} element, within the \textit{<pages>} tag, which includes dimensions and other metadata. It serves as a container for finer annotations like panels, text, characters, and speech balloons. These are described using polygonal coordinates to define their spatial positioning on the page. For instance, \textit{<panel>} and \textit{<text>} elements use a four-point polygon to delineate their boundaries, while \textit{<balloon>} elements can utilize polygons with multiple points for irregular shapes.
Moreover, UCA accommodates annotations for character reidentification and speaker identification within the comic pages, fostering more complex tasks. Elements like \textit{<characters>} belonging to the \textit{<book>} contain the character ID and names which are then used in the character annotation within the pages. Similarly, \textit{<link\_sbsc>} provides detailed annotations of the textual interactions between characters, enabling dialog tasks.
The UCA format's comprehensive approach allows researchers to meticulously annotate and analyze the complex interplay of visual and textual elements in comics. By standardizing comic book annotations, UCA aims to improve the accessibility and comparability of comic research, providing a robust foundation for advancing the field.

\begin{figure}[t]
    \centering
    \begin{minipage}[b]{0.4\linewidth}
        \centering
        \includegraphics[width=\linewidth]{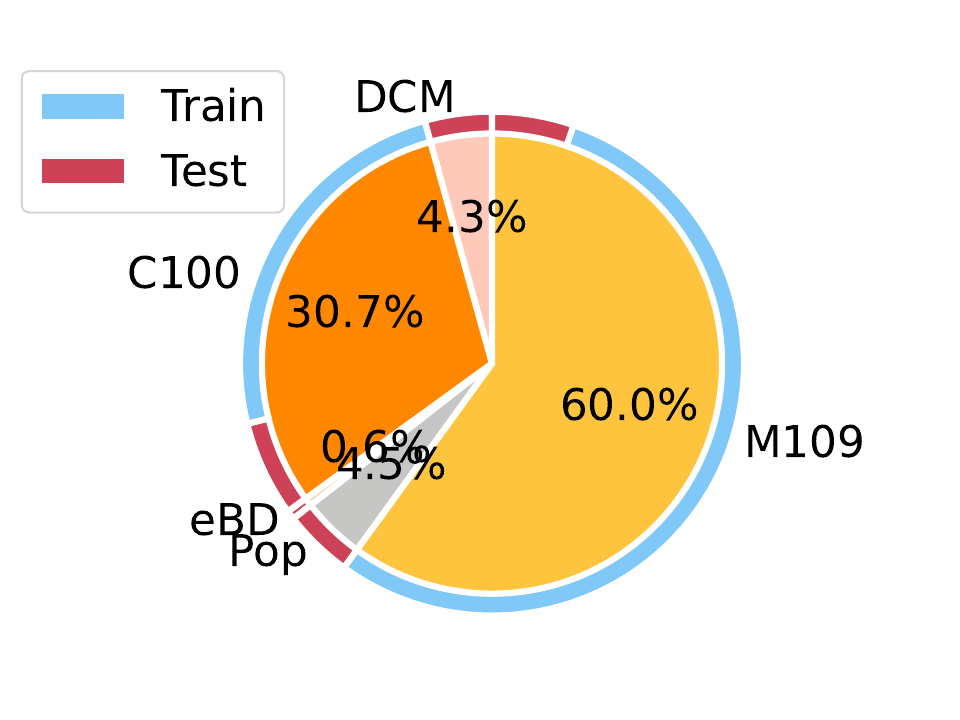}
    \caption{Dataset composition.}
    \label{fig:statistics}
    \end{minipage}
    \begin{minipage}[b]{0.58\linewidth}
        \centering
        \setlength{\tabcolsep}{4pt}
        \renewcommand{\arraystretch}{1.2} 
        \begin{tabular}{c|ccccc}
             Split & DCM & C100 & eBD & M109 &  Pop    \\
             \cline{1-6}
             Train & - & 4336 & - & 9675 & -  \\
             Test & 762 & 1086 & 100  & 927 & 789\\
             \cline{1-6}
             tot & 762 & 5422 & 100 & 10602  & 789  \\
        \end{tabular}
        \vspace{20pt}
    \captionof{table}{Details of images per datasets.}
    \label{tab:statistics}
    \end{minipage}
\end{figure}

\begin{figure}[t]
    \begin{subfigure}{0.3\linewidth}
        \centering
        \vspace{2mm}
        \includegraphics[height=4.5cm]{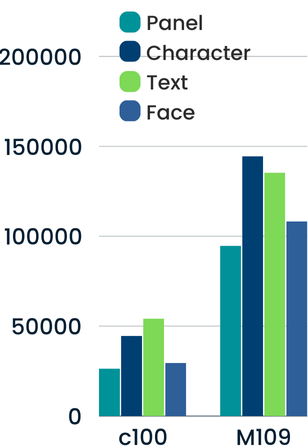}
        \caption{Training-set}
    \end{subfigure}
    \begin{subfigure}{0.6\linewidth}
        \centering
        \includegraphics[height=4.2cm]{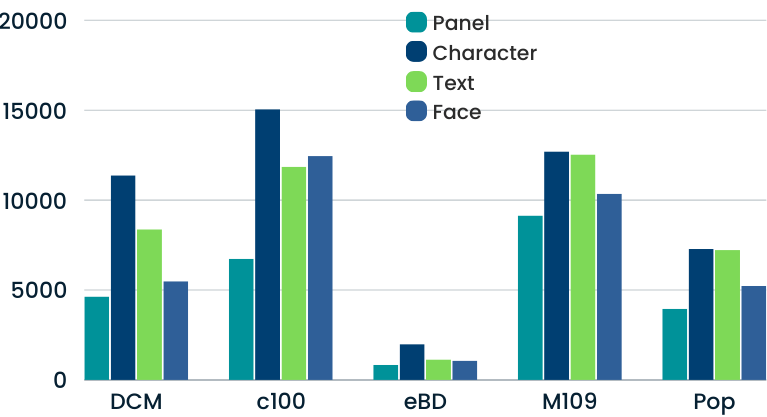}
        \caption{Test-set}
    \end{subfigure}
    \caption{Number of annotation types per dataset.}
    \label{fig:anns-statistics}
\end{figure}

\subsection{Datasets and Annotation Quality}
\label{meth:supp-datasets}
As presented in our collection of reported benchmarks in Table \ref{tab:previous-benchmarks}, various datasets have been proposed so far. However, many of them are not available and, for others, the data and annotation quality are not sufficient to be considered in this study. Therefore, we selected 4 datasets.

\paragraph{\textbf{eBDtheque (eBD).}} The eBDtheque\footnote{\href{http://ebdtheque.univ-lr.fr/registration}{http://ebdtheque.univ-lr.fr/registration}} \cite{guerin_ebdtheque_2013} is a collection of 100 pages over 20 books, mainly in French comics with also some English and Japanese percentage. Rigaud et al. \cite{rigaud_segmentation_2016} provide annotations by domain experts for 850 panels, 1092 balloons, 1550 characters, and 4691 text lines.

\paragraph{\textbf{DCM772 (DCM).}} The DCM772\footnote{\href{https://git.univ-lr.fr/crigau02/dcm_dataset}{https://git.univ-lr.fr/crigau02/dcm\_dataset}} \cite{nguyen_digital_2018} contains 772 page images from 27 golden-age comic books from Digital Comic Museum. Annotations contain panels, characters, and face bounding boxes.

\paragraph{\textbf{Manga109 (M109).}} The Manga109\footnote{\href{http://www.manga109.org/index_en.php}{http://www.manga109.org/index\_en.php}} \cite{fujimoto_manga109_2016} dataset contains 109 manga volumes from 93 different authors. This dataset has been further extended by COO \cite{baek_coo_2022} which added onomatopoeias polygons annotations and links among truncated onomatopoeias. Another set of annotations comes from Manga-Dialog \cite{li_manga109dialog_2023} which added a text-character link for (almost) every speaking text box.


\paragraph{\textbf{PopManga (Pop).}} The PopManga\footnote{\href{https://github.com/ragavsachdeva/Magi/tree/main/datasets}{https://github.com/ragavsachdeva/Magi}} \cite{sachdeva_manga_2024} dataset contains English manga titles from the most popular mangas. The dataset contains two test splits: test-seen with 1100 pages and test-unseen with 980 pages, which correspond to pages of books that the Magi model has seen and has not seen during training, respectively. In this study, we only consider the unseen split, namely ``pop'' across the images and tables reported.

\paragraph{}
A notable challenge in our study is the variability in annotation quality across diverse datasets. For instance, Manga109 \cite{fujimoto_manga109_2016} is equipped with high-quality annotations for object detection categories such as panels, characters, text, faces, and onomatopoeia, and it provides additional metadata like character names and text-character links that are supported by our data format. In contrast, the COMICS \cite{iyyer_amazing_2017} dataset offers only pseudo-labels generated by a YOLO model fine-tuned on a limited set of 500 examples for panel and text detection, resulting in annotations that lack the precision required for dependable usage. The more recently developed PopManga dataset \cite{sachdeva_manga_2024} includes comprehensive annotations for panel, character, and text detection, and it establishes links between text and characters. Given the complexity of tasks such as transcribing dialogue from comic pages, it is essential to assess models against a test data pool that is not only diverse and substantial but also of high quality. To address these needs, we have harmonized the annotations across datasets, specifically targeting object detection classes like panels, characters, text, and faces. Figure \ref{fig:statistics} details the statistics of the proposed \textit{Comics Datasets} unification, which is composed, in the test split, of an equal mix of American-style and Manga-style comics, with a minor inclusion of French comics. Moreover, the dataset's multilingual nature is reflected in Manga109's inclusion of Japanese comics, while PopManga, DCM and Comics100 feature English comics, and eBDtheque, though smaller in scale, includes comics in English, Japanese, and French. In Figure \ref{fig:anns-statistics} are provided comprehensive annotation counts per dataset and per split.


\subsection{Evaluation}
\label{meth:evaluation}
The main goal of this work is to be able to properly evaluate models on common settings. For this, we provide an evaluation system within our framework which relies on cocotools
, a set of common evaluation functions used across various projects among which \textit{``torchmetrics''} \footnote{\href{https://github.com/Lightning-AI/torchmetrics}{https://github.com/Lightning-AI/torchmetrics}}, \textit{``ultralytics''} \footnote{\href{https://github.com/ultralytics/ultralytics}{https://github.com/ultralytics/ultralytics}} and more.
By using the CoMix framework converter, one can obtain a JSON COCO-format file from XML, YOLO, or any other supported format. With the COCO-format predictions, we can evaluate per-class and global detection metrics such as precision at different $IoU$ thresholds $\{.50, .50-.95\}$, and recall with gradually increasing the number of objects detected $\{1, 10, 100\}$, considering single classes $AP, AR$, or over multiple classes $mAP, mAR$. Generally, everything starts with the two common metrics: precision and recall. Then, average precision (per class) and mean average precision (over classes) are calculated. Finally, these metrics are calculated at varying thresholds of IoU.

\textbf{Intersection over Union (IoU)} is a metric used to evaluate the accuracy of a single prediction, and measures the overlap between the predicted bounding box and the ground truth bounding box:
\begin{equation*}
    \text{\textit{IoU}} = \frac{\text{area of overlap}}{\text{area of union}}
\end{equation*}
The IoU threshold is crucial in determining whether a detection is considered a true positive. It plays a significant role in the evaluation of both the spatial accuracy of the bounding box and the correctness of the object classification.

\textbf{Precision} measures the accuracy of the positive predictions made by the model, defined as the proportion of predicted positive detections that were both correctly classified and met a minimum IoU threshold with the ground truth:
\begin{equation*}
    \text{\textit{Precision}} = \frac{\text{TP}}{\text{TP} + \text{FP}}
\end{equation*}
where TP represents true positives (correct class and sufficient IoU overlap) and FP represents false positives (incorrect class or insufficient IoU overlap).

\textbf{Recall} measures the model's ability to detect all relevant instances in the dataset, considering correct classification and IoU:
\begin{equation*}
    \text{\textit{Recall}} = \frac{\text{TP}}{\text{TP} + \text{FN}}
\end{equation*}
where FN represents false negatives (missed detections or incorrect classifications).

\textbf{Average Precision (AP)} quantifies the trade-off between precision and recall at various decision thresholds. AP is calculated as the area under the precision-recall curve (PR curve). The PR curve is derived by adjusting the decision threshold on the detection scores provided by the model, considering each class separately:
\begin{equation*}
    \text{\textit{AP}} = \sum_{n} (R_n - R_{n-1}) P_n
\end{equation*}
where $P_n$ and $R_n$ are the precision and recall at the nth threshold.

The \textbf{Mean Average Precision (mAP)} is the mean of the AP scores calculated for all classes:
\begin{equation*}
    \text{\textit{mAP}} = \frac{1}{N} \sum_{i=1}^N \text{AP}_i
\end{equation*}
where $N$ is the number of classes. The mAP can be calculated at different IoU thresholds to provide insights into the model's performance across various levels of localization strictness.

In some benchmarks, such as those used in competitions like COCO\footnote{\href{https://cocodataset.org/\#detection-eval}{https://cocodataset.org/\#detection-eval}} (Common Objects in Context), the $mAP$ is calculated at multiple IoU thresholds (e.g., 0.5 to 0.95 in steps of 0.05). This involves computing the precision and recall at each of these IoU levels to get a more nuanced understanding of the model's performance across different criteria of spatial accuracy.

\section{Benchmarking}
\label{benchmarking}

The selected models for benchmarking comprehend convolution- and transformer-based models finetuned on comics data or in a zero-shot setting. Among the zero-shot, GroundingDino \cite{liu_grounding_2023} is our main choice. This model is trained for the task of open-set object detection, an extended version of object detection with natural language object classes. 

We used GroundingDino with different class names to detect the four classes of objects: panels, characters, text, and faces. For each of these objects, we have merged results obtained with similar object classes that could help the model to resemble comics elements, such as: (panels) ``comics panels'', ``manga panels'', ``frames'', ``windows''; (characters) ``characters'', ``comics characters'', ``person'', ``girl'', ``woman'', ``man'', ``animal''; (text) ``text box'', ``text'', ``handwriting''; and finally (face) ``face'', ``character face'', ``animal face'', ``head'', ``face with nose and mouth'', ``person's face''. All the other models, instead, have been specifically trained on comics data. 

DASS is a convolution-based model (YOLOX) trained in a self-supervised approach distilling a teacher network with the OHEM loss. The model comes with three weight versions: dcm, manga, and mix. The three different model instances have been fine-tuned on the corresponding datasets, which should correspond to higher scores in the same distribution of comic style.

Additionally, we have selected three model classes that have been used multiple times in previous works: Faster R-CNN, SSD, and YOLO. For each of these convolution-based models, we have trained on ``comics-style'', ``manga-style'', and ``mixed comics-styles'', following \cite{topal_dassdetector_2023}. The ``comics-style'' training set is obtained by the Comics100 train split (almost 4.5k pages), which we will release in the \textit{CDF} repository as part of this work. The ``manga-style'' train set is the one provided by Manga109, which corresponds to 9k images. When a model has been trained on the ``mixed comics-style'' set, we refer to the fusion of Comics100 and Manga109 train splits. This can show us the impact of different training data distributions on performances in different comic styles. Our Faster R-CNN initialized with the available weights in ``torchvision'' with ResNet-50 backbone. We have modified the last layer, box predictor, to support our number of classes (4) instead of the default coco classes. The model has been fine-tune for 50 epochs with SGD with a \textit{learning rate} $5e^{-3}$, \textit{momentum} and \textit{weight\_decay} of $0.9$ and $5e^{-4}$, respectively. We have used a \textit{StepLR} scheduler inter-epochs and a \textit{CosineAnnealingLR} scheduler at every epoch. We have rescaled the images to $(1024\times1024)$ and applied \textit{RandomHorizontalFlip} with probability $0.5$. The SSD model, instead, has been trained using the ``mmdetection'' framework\footnote{\href{https://github.com/open-mmlab/mmdetection}{https://github.com/open-mmlab/mmdetection}}, using the default configurations for ssd300, and scaling input images to $(1024\times1024)$, as for Faster R-CNN. We have fine-tuned SSD on the three datasets, all with the same configuration. Finally, the YOLOv8 model has been fine-tuned from the pre-trained weights available in ``ultralytics'', following the default yolov8x configuration. All these experiments were carried out with a batch size of 32, in a single A40 GPU.

The last model we have tested on our \textit{Comics Datasets} collection is the recent Magi \cite{sachdeva_manga_2024}, a transformer-based model with DeTr backbone and two MLP heads for character re-identification and text-character linking, inspired from \cite{shit_relationformer_2022}. Authors pre-trained Magi on GroundingDINO annotated Mangadex\footnote{\href{https://mangadex.org/}{https://mangadex.org/}} noisy dataset and fine-tuned on an \textit{un-available} manually annotated popmanga dev-set.

The detection results are reported in Tables \ref{tab:bench:all}, where the used metric is the mean Average Precision (mAP) averaged across the four classes considered. However, some model (DASS and Magi) only considers two and three classes, respectively. Thus, we provide the mAP score averaging only detectable classes in Table \ref{tab:bench:all-filtered}.
In the Appendix, we reported mAP detailed for panels (table \ref{tab:bench:panel}), characters (table \ref{tab:bench:character}), faces (table \ref{tab:bench:face}) and text (\ref{tab:bench:text}). Other metrics are reported in the repository, across which the mAP averaged across the 0.5\%-0.95\% IoU, Average Recall at 1 (AR@1) and 100 (AR@100), which are common metrics for object detection tasks.

In the tables, we emphasize that GroundingDino is used in a zero-shot setting, not specifically trained for the detection of our classes; the Faster R-CNN, SSD, and YOLOv8 correspond to our fine-tuned models; while DASS and Magi are the available models specifically tailored for comics and/or manga. An additional note is that DASS presents in its training a subset of the DCM dataset split. This is because we consider DCM (the whole dataset) as a test-set split. The choice of considering the whole DCM for testing in our \textit{Comics Datasets} collection is due to its small size and availability of well-curated annotations.

\begin{table}
\hspace{-15pt}
\begin{minipage}[b]{0.49\textwidth}
\centering
\small
\begin{tabular}{rccccc|c}
& \textbf{DCM} & \textbf{c100} & \textbf{eBD} & \textbf{M109} & \textbf{Pop} & \textit{avg} \\
\hline
G.Dino & 63,4 & 62,5 & 56,9 & 61,8 & 73,7 & 64,7 \\
R-CNN & {\ul 86,3} & \textbf{88,9} & {\ul 65,4} & 64,9 & {\ul 77,6} & \textbf{79,5} \\
SSD & 12,1 & 9,1 & 28,4 & 34,6 & 4,6 & 15,6 \\
YOLO & 81,4 & {\ul 75,0} & \textbf{67,0} & \textbf{76,8} & 64,5 & 74,3 \\
DASS & - & - & - & - & - & - \\
Magi & \textbf{89,0} & 73,9 & 62,1 & {\ul 65,3} & \textbf{92,8} & {\ul 78,5} \\
\hline
\end{tabular}
\vspace{3mm}
\caption{\textbf{Panel} detection.}
\label{tab:bench:panel}
\end{minipage}
\hspace{15pt}
\begin{minipage}[b]{0.49\textwidth}
\centering
\small
\begin{tabular}{rccccc|c}
 & \textbf{DCM} & \textbf{c100} & \textbf{eBD} & \textbf{M109} & \textbf{Pop} & \textit{avg} \\
\hline
G.Dino & 57,7 & 62,1 & 40,1 & 25,3 & 46,0 & 48,2 \\
R-CNN & 50,6 & 61,0 & 34,7 & 4,7 & 50,9 & 42,2 \\
SSD & 52,4 & 54,1 & 39,5 & {\ul 55,8} & 32,6 & 49,3 \\
YOLO & 45,6 & 55,4 & 30,1 & 9,4 & 42,0 & 38,6 \\
DASS & \textbf{75,1} & {\ul 76,0} & \textbf{60,9} & \textbf{84,4} & {\ul 70,5} & \textbf{76,3} \\
Magi & {\ul 71,8} & \textbf{76,7} & {\ul 56,6} & 50,4 & \textbf{79,7} & {\ul 69,3} \\
\hline
\end{tabular}
\vspace{3mm}
\caption{\textbf{Characters} detection.}
\label{tab:bench:character}
\end{minipage}
\end{table}
\begin{table}
\hspace{-15pt}
\begin{minipage}[b]{0.49\textwidth}
\centering
\begin{tabular}{rrrrrr|c}
 & \multicolumn{1}{c}{\textbf{DCM}} & \multicolumn{1}{c}{\textbf{c100}} & \multicolumn{1}{c}{\textbf{eBD}} & \multicolumn{1}{c}{\textbf{M109}} & \multicolumn{1}{c|}{\textbf{Pop}} & \textit{avg} \\
 \hline
G.Dino & {\ul 66,5} & 58,9 & {\ul 37,3} & 38,1 & 62,0 & 55,4 \\
R-CNN & 43,0 & 38,9 & 20,7 & 8,7 & 43,0 & 32,7 \\
SSD & 60,1 & {\ul 60,0} & 30,9 & {\ul 76,4} & {\ul 75,4} & {\ul 66,5} \\
YOLO & 43,1 & 48,8 & 20,6 & 16,2 & 42,1 & 37,5 \\
DASS & \multicolumn{1}{c}{\textbf{78,8}} & \multicolumn{1}{c}{\textbf{62,7}} & \multicolumn{1}{c}{\textbf{61,1}} & \multicolumn{1}{c}{\textbf{87,8}} & \multicolumn{1}{c|}{\textbf{78,0}} & \textbf{75,3} \\
Magi & \multicolumn{1}{c}{-} & \multicolumn{1}{c}{-} & \multicolumn{1}{c}{-} & \multicolumn{1}{c}{-} & \multicolumn{1}{c|}{-} & - \\
\hline
\end{tabular}
\vspace{3mm}
\caption{\textbf{Face} detection.}
\label{tab:bench:face}
\end{minipage}
\hspace{15pt}
\begin{minipage}[b]{0.49\textwidth}
\centering
\begin{tabular}{rrrrrr|c}
 & \multicolumn{1}{c}{\textbf{DCM}} & \multicolumn{1}{c}{\textbf{c100}} & \multicolumn{1}{c}{\textbf{eBD}} & \multicolumn{1}{c}{\textbf{M109}} & \multicolumn{1}{c|}{\textbf{Pop}} & \textit{avg} \\
 \hline
G.Dino & 20,7 & 23,0 & 17,8 & 9,9 & 27,6 & 20,1 \\
R-CNN & 64,2 & \textbf{83,1} & {\ul 41,9} & 14,4 & {\ul 48,5} & 54,0 \\
SSD & 58,5 & 70,2 & 38,5 & \textbf{70,8} & 31,7 & {\ul 59,1} \\
YOLO & {\ul 68,3} & 73,0 & 38,7 & 42,2 & 12,7 & 50,9 \\
DASS & \multicolumn{1}{c}{-} & \multicolumn{1}{c}{-} & \multicolumn{1}{c}{-} & \multicolumn{1}{c}{-} & \multicolumn{1}{c|}{-} & - \\
Magi & \multicolumn{1}{c}{\textbf{84,0}} & \multicolumn{1}{c}{{\ul 77,9}} & \multicolumn{1}{c}{\textbf{73,6}} & \multicolumn{1}{c}{{\ul 49,2}} & \multicolumn{1}{c|}{\textbf{93,4}} & \textbf{75,2} \\
\hline
\end{tabular}
\vspace{3mm}
\caption{\textbf{Text} detection.}
\label{tab:bench:text}
\end{minipage}
\end{table}

\begin{table}
\hspace{-15pt}
\begin{minipage}[t]{0.49\textwidth}
\centering
\begin{tabular}{rccccc|c}
 & \textbf{DCM} & \textbf{c100} & \textbf{eBD} & \textbf{M109} & \textbf{Pop} & \textit{avg} \\
 \hline
G.Dino & 34,7 & 36,9 & 25,3 & 27,0 & 36,8 & 33,7 \\
R-CNN & {\ul 40,7} & \textbf{58,2} & {\ul 27,1} & 18,5 & {\ul 44,2} & {\ul 41,1} \\
SSD & 30,5 & 33,3 & 22,9 & \textbf{47,5} & 17,2 & 32,6 \\
YOLO & 39,7 & 50,8 & 26,1 & 28,9 & 29,8 & 38,1 \\
DASS & 25,7 & 19,0 & 20,3 & {\ul 34,4} & 17,6 & 23,9 \\
Magi & \textbf{40,8} & {\ul 57,1} & \textbf{32,1} & 33,0 & \textbf{66,5} & \textbf{49,2} \\
\hline
\end{tabular}
\vspace{2mm}
\caption{Averaged mAP across \textbf{all classes}.}
\label{tab:bench:all}
\end{minipage}
\hspace{15pt}
\begin{minipage}[t]{0.49\linewidth}
\centering
\begin{tabular}{rrrrrr|r}
 & \multicolumn{1}{c}{\textbf{DCM}} & \multicolumn{1}{c}{\textbf{c100}} & \multicolumn{1}{c}{\textbf{eBD}} & \multicolumn{1}{c}{\textbf{M109}} & \multicolumn{1}{c|}{\textbf{Pop}} & \multicolumn{1}{c}{\textit{avg}} \\
 \hline
G.Dino & 52,1 & 51,1 & 38,0 & 33,8 & 52,0 & 46,9 \\
R-CNN & 61,0 & 66,0 & 40,7 & 23,1 & 59,0 & 52,3 \\
SSD & 45,8 & 44,5 & 34,3 & {\ul 59,4} & 23,0 & 43,7 \\
YOLO & 59,6 & 63,0 & 39,1 & 36,1 & 54,1 & 53,2 \\
DASS & {\ul 77,0} & {\ul 67,6} & {\ul 61,0} & \textbf{86,1} & {\ul 73,3} & \textbf{75,1} \\
Magi & \textbf{81,6} & \textbf{75,6} & \textbf{64,1} & 54,9 & \textbf{88,0} & {\ul 74,0} \\
\hline
\end{tabular}
\vspace{2mm}
\caption{Averaged mAP on \textbf{detected classes}.}
\label{tab:bench:all-filtered}
\end{minipage}
\end{table}

In these tables, we present several significant values for mean average precision. Notably, the most revealing insights arise from the analysis of failures. The SSD model demonstrates the ability to detect characters, faces, and text with around 60\% average precision (AP.5), yet it significantly underperforms in identifying panels across various datasets. This issue does not stem from an excess of bounding boxes, as indicated by the low recall rates (recall and other metrics are reported in the repository). It is more likely related to the specific characteristics of the datasets employed for pertaining. For instance, the comic datasets (referred to in the tables as c100) are generated through automatic annotation from multiple sources. Panels are identified using a Yolov8 model fine-tuned on 500 pages solely annotated for panels as detailed by \cite{iyyer_amazing_2017}. Character and face detections are enhanced by the DASS-dcm, which is fine-tuned on DCM books. Textboxes are identified using text extracted by Amazon Textract as reported in \cite{vivoli_multimodal_2024}, where lines within the same balloon are combined. Therefore, we have already established various upper-bounds for different classes. Additional unexpected failures include Faster R-CNN’s performance on Japanese Manga109 images and Yolov8 (m109 version) in detecting characters, text, and panels outside its domain. With Faster R-CNN, the model performs moderately for panels but is less effective with other classes, detecting only sporadic texts and faces. Conversely, Yolov8, specifically tuned for Manga109, shows average performance on manga-style comics within its domain but falls short on other comic styles.

\section{Conclusion}
\label{conclusion}
In conclusion, our paper introduces the \textit{Comics Datasets Framework}, designed to facilitate the management of datasets with restricted access through conversion scripts that transform custom annotations and folder structures into a Unified Comics Annotation. This framework not only corrects and extends annotations for existing datasets but also introduces new annotations for comic styles that are less represented compared to manga styles. Additionally, a primary contribution of our work is establishing comparable and replicable experiments through common setting baselines and making models available. We have constructed the largest test set of various comic styles, namely \textit{Comics Datasets}, by enhancing and integrating existing annotations, and we provide a standardized evaluation setting to enable fair comparison of both existing and new methods. We demonstrate how to effectively benchmark both existing and new comic detection models, providing access to the model weights and code to ensure reproducibility. We believe this initiative is crucial for clarifying the comics research landscape and enabling consistent reproducibility and benchmarking.

\newpage

%
%
%
\bibliographystyle{splncs04}
\bibliography{survey}

\end{document}